\documentclass[10pt,twocolumn,letterpaper]{article}

\usepackage{cvpr}
\usepackage{times}
\usepackage{epsfig}
\usepackage{graphicx}
\usepackage{amsmath}
\usepackage{amssymb}

\usepackage{wrapfig}
\usepackage{hhline}
\usepackage{booktabs}
\usepackage{commath}
\usepackage{shortbold}

\usepackage[pagebackref=true,breaklinks=true,letterpaper=true,colorlinks,bookmarks=false]{hyperref}

\cvprfinalcopy % *** Uncomment this line for the final submission

\def\cvprPaperID{6332} % *** Enter the CVPR Paper ID here

\ifcvprfinal\fi
\begin{document}

%%%%%%%%% TITLE
\title{Tangent-Normal Adversarial Regularization for Semi-supervised Learning}

\author{Bing Yu\footnotemark[1] \& Jingfeng Wu\footnotemark[1]\thanks{Equal contributions.} \\
School of Mathematical Sciences\\
Peking University\\
Beijing, China \\
\tt{\small\{byu,pkuwjf\}@pku.edu.cn} \\
\and
Jinwen Ma \\
School of Mathematical Sciences\\
Peking University\\
Beijing, China \\
\tt{\small\{jwma\}@math.pku.edu.cn} \\
\and
Zhanxing Zhu\footnotemark[2]\thanks{Corresponding author.} \\
Center for Data Science Peking University \& \\
Beijing Institute of Big Data Research\\
Beijing, China \\
\tt{\small\{zhanxing.zhu\}@pku.edu.cn} \\
}

% \author{First Author\\
% Institution1\\
% Institution1 address\\
% {\tt\small firstauthor@i1.org}
% % For a paper whose authors are all at the same institution,
% % omit the following lines up until the closing ``}''.
% % Additional authors and addresses can be added with ``\and'',
% % just like the second author.
% % To save space, use either the email address or home page, not both
% \and
% Second Author\\
% Institution2\\
% First line of institution2 address\\
% {\tt\small secondauthor@i2.org}
% }

\maketitle
%\thispagestyle{empty}

%%%%%%%%% ABSTRACT
\begin{abstract}
    %Though developed years ago, virtual adversarial training (VAT) introduced by Miyato et.al. is still one of the state-of-the-art methods for semi-supervised learning.
    Compared with standard supervised learning, the key difficulty in semi-supervised learning is how to make full use of the unlabeled data. A recently proposed method, virtual adversarial training (VAT),  smartly performs adversarial training without label information to  impose a local smoothness on the classifier, which is especially beneficial to semi-supervised learning.
    In this work, we propose \textbf{tangent-normal adversarial regularization} (TNAR) as an extension of VAT by taking the data manifold into consideration.
    The proposed TNAR is composed by two complementary parts, the tangent adversarial regularization (TAR) and the normal adversarial regularization (NAR).
    In TAR, VAT is applied along the tangent space of the data manifold, aiming to enforce local invariance of the classifier on the manifold,
    while in NAR, VAT is performed on the normal space orthogonal to the tangent space, intending to impose robustness on the classifier against the noise causing the observed data deviating from the underlying data manifold.
    Demonstrated by experiments on both artificial and practical datasets, our proposed TAR and NAR complement with each other, and jointly outperforms other state-of-the-art methods for semi-supervised learning.
\end{abstract}

%%%%%%%%% BODY TEXT
\section{Introduction}
\label{sec:intro}
The main challenge in semi-supervised learning (SSL) is how to utilize the large amount of the unlabeled data to obtain useful information, benefiting the supervised learning on the relatively insufficient amount of labeled data.
For this purpose, one of the important line of research focuses on the manifold assumption on the data distribution,
i.e., the observed data is distributed on a low dimensional manifold that could be characterized using the large amount of the unlabeled data, and aims to learn a proper classifier based on the data manifold~\cite{belkin2006manifold,rifai2011manifold,niyogi2013manifold,kumar2017semi,lecouat2018semi}. 
Following this stream, we sort out three reasonable assumptions to motivate our idea for semi-supervised learning:
\begin{description}
    \item[The manifold assumption] The observed data $x$ presented in high dimensional space $\Rbb^D$ is with high probability concentrated in the vicinity of some underlying manifold with much lower dimensionality~\cite{cayton2005algorithms,narayanan2010sample,chapelle2009semi,rifai2011manifold}, denoted as $\Mcal\cong \Rbb^d$.
    %  and further assume that the classification task concerned relies and only relies on $\Mcal$~\cite{rifai2011manifold}.
    \item[The noisy observation assumption] The observed data $x$ can be decomposed into two parts as $x=x_0+n$, where $x_0$ is exactly supported on the underlying manifold $\Mcal$ and $n$ is some noise independent of $x_0$~\cite{bengio2013generalized,rasmus2015semi}.
    % With the assumption that the classifier only depends on the underlying manifold $\Mcal$, the noise part might have undesired influence on the learning of the classifier.
    \item[The semi-supervised learning assumption] If two points $x_1, x_2 \in \Mcal$ are close in manifold distance, then the conditional probability $p(y|x_1)$ and $p(y|x_2)$ are similar~\cite{belkin2006manifold,rifai2011manifold,niyogi2013manifold}. In other words, the true classifier, or the true condition distribution $p(y|X)$ varies smoothly along the underlying manifold $\Mcal$.
\end{description}

According to the three assumptions, the best classifier we aim to obtain should be 
1) smooth along the data manifold;
2) robust to the off-manifold noise.
Hence it is natural to formulate a loss function~\cite{belkin2006manifold,kumar2017semi} for SSL as,
\begin{equation}
    L_{\text{ssl}} := L_{\text{supervised}} + \Rcal_{\text{manifold}} + \Rcal_{\text{noise}},
    \label{eq:loss_ssl}
\end{equation}
where the first term in Eq.~(\ref{eq:loss_ssl}) is the supervised learning loss,
the second term penalize the manifold smoothness of the classifier,
and the third term smooths the classifier so that it is robust to noise, respectively. 
While the supervised learning loss $L_{\text{supervised}}$ concerning the labeled data is standard, the key ingredient lies on how to design $\Rcal_{\text{manifold}}$ and $\Rcal_{\text{noise}}$ smartly to 1) be effective for inducing the desired smoothness on the classifier, 2) be efficient for optimization, and 3) make full use of the unlabeled data.
 
Existing works construct $\Rcal_{\text{manifold}}$ based on the Jacobian,
for instance, \emph{tangent propagation}~\cite{simard1998transformation,kumar2017semi}
\begin{equation}
    \Rcal_{\text{manifold}} = \Ebb_{x\sim p(x)} \sum_{v\in T_x \Mcal} \norm{(J_x f) \cdot v},
    \label{eq:tangent-prop}
\end{equation}
and \emph{manifold Laplacian norm}~\cite{belkin2006manifold,lecouat2018semi,qi2018global} 
\begin{equation}
    \begin{aligned}
        \Rcal_{\text{manifold}} & = \int_{x\in \Mcal} \norm{\down_{\Mcal} f(x)} \dif p(x) \\
        & \approx \Ebb_{z\sim p(z)} \norm{\down_{\Mcal} f(g(z))} \\
        & \approx \Ebb_{z\sim p(z)} \norm{J_z f(g(z))},
    \end{aligned}
    \label{eq:laplacian}
\end{equation}
where $J$ is the Jacobian, $f$ is the classifier, $T_x\Mcal$ is the tangent space of the data manifold and $x=g(z)$ is the manifold representation of data.
They regularize the manifold smoothness of the classifier under the sense of the norm of its Jacobian along the data manifold. The typical choice of $\Rcal_{\text{noise}}$ is in a corresponding form as $\Rcal_{\text{manifold}}$ except the Jacobian to penalize is with respect to the observation space other than the tangent space~\cite{belkin2006manifold,kumar2017semi}. 

On the other hand, inspired by adversarial training~\cite{goodfellow2014explaining}, virtual adversarial training (VAT)~\cite{miyato2017virtual,miyato2016adversarial} was proposed for SSL, not relying on the label information.
Unlike the smoothness induced by $L_p$-norm of the Jacobian,
VAT leads to the robustness of classifier by involving virtual adversarial examples,
thus inducing a new local smoothness of the classifier.
Empirical results~\cite{miyato2017virtual,oliver2018realistic} show that VAT achieves state-of-the-art performance for SSL tasks, demonstrating the superiority of the smoothness imposed by virtual adversarial training.

\begin{figure}
    \centering
    \includegraphics[width=0.5\linewidth]{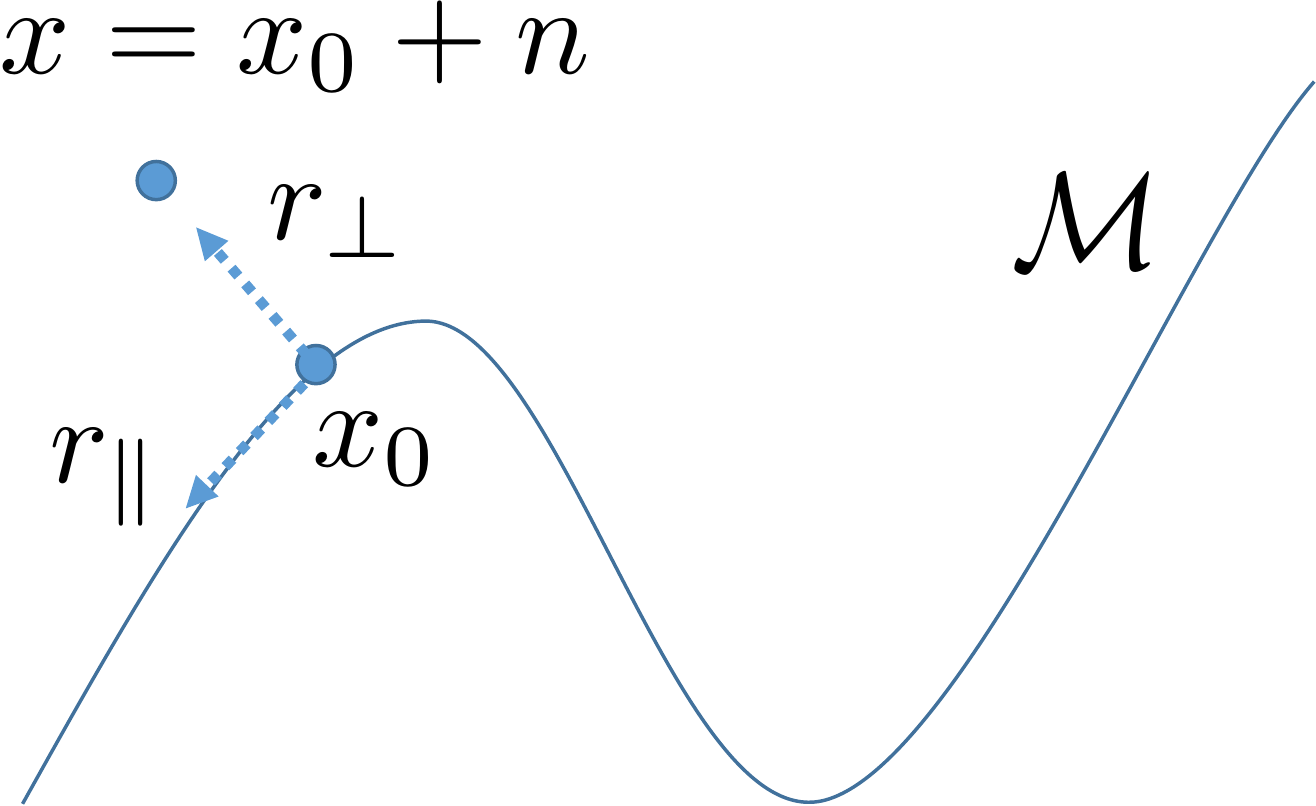}
    \caption{Illustration for tangent-normal adversarial regularization. $x=x_0+n$ is the observed data, where $x_0$ is exactly supported on the underlying manifold $\Mcal$ and $n$ is the noise independent of $x_0$. $r_{\parallel}$ is the adversarial perturbation along the tangent space to induce invariance of the classifier on manifold; $r_{\bot}$ is the adversarial perturbation along the normal space to impose robustness on the classifier against noise $n$.}
    \label{fig:decomposition}
\end{figure}

Encouraged by the effectiveness of VAT, we propose to construct manifold regularizer based on VAT, instead of the $L_p$-norm of the Jacobian.
Concretely, we propose \emph{tangent adversarial regularization} (TAR) by performing VAT along the tangent space of the data manifold, and \emph{normal adversarial regularization} (NAR) by applying VAT orthogonal to the tangent space of the data manifold, which are intuitively demonstrated in Figure~\ref{fig:decomposition}.
TAR enforces the local smoothness of the classifier along the underlying manifold, while NAR imposes robustness on the classifier against the noise carried in the observed data.
The two terms, complementing with each other, establish our proposed approach \emph{tangent-normal adversarial regularization} (TNAR).

To realize TNAR, we have two challenges to conquer: 1) how to estimate the underlying manifold and 2) how to efficiently perform TNAR.
For the first issue, we take advantage of the generative models equipped with an extra encoder, to characterize the coordinate chart of manifold~\cite{kumar2017semi,lecouat2018semi,qi2018global}. More specifically, in this work we choose variational autoendoer (VAE)~\cite{kingma2013auto} and localized GAN~\cite{qi2018global} to estimate the underlying manifold from data.
For the second problem, we further extend the techniques introduced in ~\cite{miyato2017virtual} with some sophisticatedly  designed auxiliary functions, implementing VAT restricted in tangent space (TAR) and normal space (NAR)  efficiently. The details are elaborated in Section~\ref{sec:method}.

The remaining of the paper is organized as follows. In Section~\ref{sec:background} we introduce VAT and two generative models as the background of TNAR. Based on that, we elaborate about the technical details of TNAR in Section~\ref{sec:method}. In Section~\ref{sec:compare} we compare TNAR with other related approaches and analyze the advantages of TNAR over VAT and other manifold-based regularization. Various experiments are conducted for demonstrating the effectiveness of TNAR in Section~\ref{sec:exp}. And in Section~\ref{sec:discuss} and Section~\ref{sec:conclude}, we discuss an existing problem about TNAR for future exploration and conclude the paper. 

\section{Background}
\label{sec:background}
\subsection{Notations}
The labeled and unlabeled dataset are denoted as $\Dcal_l = \{(x_l, y_l)\}$ and $\Dcal_{ul} = \{x_{ul}\}$ respectively, thus $\Dcal := \Dcal_l \cup \Dcal_{ul}$ is the full dataset. The output of classification model is written as $p(y|x, \theta)$, with $\theta$ being the model parameters to be trained. $\ell(\cdot, \cdot)$ represents the supervised loss function.
For data example, the observed space $\Rbb^D$ and the underlying manifold is $\Mcal$.
The decoder (generator) and the encoder are denoted as $g$ and $h$ respectively, which form the coordinate chart of manifold together. If not stated otherwise, we always assume $x$ and $z$ correspond to the coordinate of the same data point in observed space $\Rbb^D$ and on manifold $\Mcal$, i.e., $g(z)=x$ and $h(x)=z$.
The tangent space of $\Mcal$ at point $x$ is $T_x \Mcal = J_z g(\Rbb^d) \cong \Rbb^d$, where $J_z g$ is the Jacobian of $g$ at point $z$. $T_x \Mcal$ is also the span of the columns of $J_z g$. We use $J$ to represent the Jacobian when there is no ambiguity.

The perturbation in the observed space $\Rbb^D$ is denoted as $r \in \Rbb^D$, while the perturbation on the manifold representation is denoted as $\eta \in \Rbb^d$. Hence the perturbation on manifold is $g(z+\eta) - g(z)$. When the perturbation $\eta$ is small enough for the holding of the first order Taylor's expansion, the perturbation on manifold is approximately equal to the perturbation on its tangent space, $g(z+\eta) - g(z) \approx J \cdot \eta \in T_x \Mcal$.
Therefore we say a perturbation $r\in \Rbb^D$ is actually on manifold, if there is a perturbation $\eta\in \Rbb^d$, such that $r = J \cdot \eta$.

\subsection{Virtual adversarial training}
VAT~\cite{miyato2017virtual} is an effective regularization method for SSL.
The virtual adversarial loss introduced in VAT is defined by the robustness of the classifier against local perturbation in the input space $\Rbb^D$. Hence VAT imposes a kind of smoothness condition on the classifier.
Mathematically, the virtual adversarial loss in VAT for SSL is
\begin{equation}
\begin{aligned}
    L(\Dcal_l, \Dcal_{ul}, \theta) := & \Ebb_{(x_l, y_l)\in \Dcal_l} \ell(y_l, p(y|x_l, \theta))\\
    & + \alpha \Ebb_{x\in\Dcal}\Rcal_{\text{vat}}(x,\theta).
\end{aligned}
\end{equation}
The VAT regularization $\Rcal_{\text{vat}}$ is defined as
\begin{equation}
    \Rcal_{\text{vat}} (x;\theta) := \max_{\norm{r}_2 \le \epsilon} dist( p(y|x,\theta), p(y|x+r,\theta) ),
\end{equation}
where $dist(\cdot, \cdot)$ is some distribution distance measure and $\epsilon$ controls the magnitude of the adversarial example.
For simplicity, define
\begin{equation}
    F(x,r,\theta):= dist(p(y|x,\theta), p(y+r,\theta)).
    \label{eq:F}
\end{equation}
Then 
$\Rcal_{\text{vat}} = \max_{\norm{r}_2 \le \epsilon}F(x,r,\theta)$.
And the so called virtual adversarial example is
$r^* := \argmax_{\norm{r}\le \epsilon} F(x,r,\theta)$.
Once we have $r^*$, the VAT loss can be optimized with the objective as
$
L(\Dcal_l, \Dcal_{ul}, \theta) = \Ebb_{(x_l, y_l)\in \Dcal_l} \ell(y_l, p(y|x_l, \theta))\\
 + \alpha\Ebb_{x\in\Dcal}F(x, r^*, \theta).
$

To obtain the virtual adversarial example $r^*$, \cite{miyato2017virtual} suggested to apply second order Taylor's expansion to $F(x,r,\theta)$ around $r=0$ as
\begin{equation}
    F(x,r,\theta)\approx \half r^T H r,
    \label{eq:vat-approx}
\end{equation}
where $H:=\nabla_r^2 F(x,r,\theta)|_{r=0}$ denotes the Hessian of $F$ with respect to $r$. The vanishing of the first two terms in Taylor's expansion occurs because that $dist(\cdot,\cdot)$ is a distance measure with minimum zero and $r=0$ is the corresponding optimal value, indicating that at $r=0$, both the value and the gradient of $F(x,r,\theta)$ are zero.
Therefore for small enough $\epsilon$, $r^* \approx \argmax_{\norm{r}_2 \le \epsilon} \half r^T H r$, which is an eigenvalue problem and the direction of $r^*$ can be solved by power iteration.
% \begin{equation}
%         r \leftarrow Hr;\qquad r \leftarrow r/\norm{r}_2.
% \end{equation}
% Note that the evaluation of matrix vector product $Hr$ is enough for power iteration, which could be efficiently computed since derivation is a linear operator and can communicate with vector multiplication.

\subsection{Generative models for data manifold}
\label{sec:manifold-model}
We take advantage of generative model with both encoder $h$ and decoder $g$ to estimate the underlying data manifold $\Mcal$ and its tangent space $T_x\Mcal$. As assumed by previous works~\cite{kumar2017semi,lecouat2018semi}, perfect generative models with both decoder and encoder can describe the data manifold, where the decoder $g(z)$ and the encoder $h(x)$ together serve as the coordinate chart of manifold $\Mcal$.
Note that the encoder is indispensable for it helps to identify the manifold coordinate $z=h(x)$ for point $x\in \Mcal$.
With the trained generative model, the tangent space is given by $T_x\Mcal = J_z g(\Rbb^d)$, or the span of the columns of $J=J_z g$.

In this work, we adopt VAE~\cite{kingma2013auto} and localized GAN~\cite{qi2018global} to learn the targeted underlying data manifold $\Mcal$ as summarized below.

\textbf{VAE}
VAE~\cite{kingma2013auto} is a well known generative model consisting of both encoder and decoder. The training of VAE is by optimizing the variational lower bound of log likelihood,
\begin{equation}
    \begin{aligned}
        \log p(x,\theta) \ge & \Ebb_{z\sim q(z|x, \theta)} \left[ \log p(x|z, \theta) \right] \\
        &- KL(q(z|x, \theta)\|p(z)).
    \end{aligned}
\end{equation}
Here $p(z)$ is the prior of hidden variable $z$, and $q(z|x,\theta)$, $p(x|z,\theta)$ models the encoder and decoder in VAE, respectively. The derivation of the lower bound with respect to $\theta$ is well defined thanks to the reparameterization trick, thus it could be optimized by gradient based method.
The lower bound could also be interpreted as a reconstruction term plus a regularization term~\cite{kingma2013auto}.
With a trained VAE, the encoder and decoder are given as $h(x) = \argmax_z q(z|x)$ and $g(z) = \argmax_x q(x|z)$ accordingly.

\textbf{Localized GAN}
Localized GAN~\cite{qi2018global} suggests to use a localized generator $G(x,z)$ to replace the global generator $g(z)$ in vanilla GAN~\cite{goodfellow2014generative}.
The key difference between localized GAN and previous generative model for manifold is that, localized GAN learns a distinguishing local coordinate chart for each point $x\in \Mcal$, which is given by $G(x,z)$, rather than one global coordinate chart.
To model the local coordinate chart in data manifold, localized GAN requires the localized generator to satisfy two more regularity conditions:
\begin{description}
    \item[locality] $G(x,0)=x$, so that $G(x,z)$ is localized around $x$;
    \item[orthogonmality] $\left(\frac{\partial G(x,z)}{\partial z}\right)^T \frac{\partial G(x,z)}{\partial z} = I$, to ensure $G(x,z)$ is non-degenerated.
\end{description}
The two conditions are achieved by the following penalty during training of localized GAN:
\begin{equation}
    \begin{aligned}
        \Rcal_{\text{localized GAN}} & :=  \mu_1 \norm{G(x,0)-x}^2 + \\
        & \mu_2 \left\|\left(\frac{\partial G(x,z)}{\partial z}\right)^T \frac{\partial G(x,z)}{\partial z}-I\right\|^2.
    \end{aligned}
\end{equation}
Since $G(x,z)$ defines a local coordinate chart for each $x$ separately, in which the latent encode of $x$ is $z=0$, there is no need for the extra encoder to provide the manifold representation of $x$.

\section{Method}
\label{sec:method}
In this section we elaborate our proposed \emph{tangent-normal adversarial regularization} (TNAR) strategy. The TNAR loss to be minimized for SSL is
\begin{equation}
    \begin{aligned}
        L(D_l, D_{ul}, \theta) :=& \Ebb_{(x_l, y_l)\in\Dcal_{l}} \ell\left(y_l, p(y|x_l, \theta)\right)\\
        &+ \alpha_1 \Ebb_{x\in\Dcal} \Rcal_{\text{tangent}}(x,\theta) \\
        &+ \alpha_2 \Ebb_{x\in\Dcal} \Rcal_{\text{normal}}(x,\theta).
    \end{aligned}
    \label{eq:loss-tangent-normal}
\end{equation}
The first term in Eq.~(\ref{eq:loss-tangent-normal}) is a common used supervised loss, e.g., negative cross entropy. $\Rcal_{\text{tangent}}$ and $\Rcal_{\text{normal}}$ is the so called \emph{tangent adversarial regularization} (TAR) and \emph{normal adversarial regularization} (NAR) accordingly,  jointly forming the proposed TNAR.
In the following section, we assume that we already have a well trained generative model for the underlying data manifold $\Mcal$, with encoder $h$ and decoder $g$, which can be obtained as described in Section~\ref{sec:manifold-model}.

\subsection{Tangent adversarial regularization}
Vanilla VAT penalizes the variety of the classifier against local perturbation in the input space $\Rbb^D$~\cite{miyato2017virtual}, which might overly regularize the classifier, since the semi-supervised learning assumption only indicates that the true conditional distribution varies smoothly along the underlying manifold $\Mcal$, but not the whole input space $\Rbb^D$~\cite{belkin2006manifold,rifai2011manifold,niyogi2013manifold}.
To avoid this shortcoming of vanilla VAT, we propose the tangent adversarial regularization (TAR), which restricts virtual adversarial training to the tangent space of the underlying manifold $T_x\Mcal$, to enforce manifold invariance property of the classifier.
\begin{equation}
    \Rcal_{\text{tangent}} (x;\theta) := \max_{\norm{r}_2 \le \epsilon, r\in T_x \Mcal = J_z g(\Rbb^d)} F(x,r,\theta),
    \label{eq:tangent-vat}
\end{equation}
where $F(x,r,\theta)$ is defined as in Eq.~(\ref{eq:F}).
To optimize Eq.~(\ref{eq:tangent-vat}), we first apply Taylor's expansion to $F(x,r,\theta)$ so that
\begin{equation}
    \Rcal_{\text{tangent}} (x;\theta) \approx \max_{\norm{r}_2 \le \epsilon,\\ r\in T_x \Mcal = J_z g(\Rbb^d)} \half r^T H r,
\end{equation}
where the notations and the derivation are as in Eq.~(\ref{eq:vat-approx}).
We further reformulate $\Rcal_{\text{tangent}}$ as
\begin{equation}
    \begin{aligned}
        \underset{r\in\Rbb^D}{\text{maximize}} &\quad \half r^T H r,\\
        \text{s.t.} &\quad \norm{r}_2\leq \epsilon, \quad r = J \eta, \quad \eta\in \Rbb^d. \\
        (J&:= J_z g\in\Rbb^{D\times d},\quad H\in\Rbb^{D\times D})    
    \end{aligned}
\end{equation}
Or equivalently,
\begin{equation}
    \begin{aligned}
        \underset{\eta\in\Rbb^d}{\text{maximize}} &\quad \half \eta^T J^T H J \eta,\\
        \text{s.t.} &\quad \eta^T J^T J \eta \le \epsilon^2.
    \end{aligned}
    \label{eq:tangent-vat-eigen}
\end{equation}
This is a classic generalized eigenvalue problem, the optimal solution $\eta^*$ of which could be obtained by power iteration and conjugate gradient (and scaling). The iteration framework is as
\begin{equation}
    \begin{aligned}
        v &\leftarrow J^T HJ \eta;\\
        \mu &\leftarrow (J^T J)^{-1}v;\\
        \eta &\leftarrow \frac{\mu}{\norm{\mu}_2}.    
    \end{aligned}
    \label{eq:power-cg}
\end{equation}
Now we elaborate the detailed implementation of each step in Eq.~(\ref{eq:power-cg}).

\subparagraph{Computing $J^T HJ \eta$.}
Note that $z=h(x), x=g(z)$.
Define
\begin{equation}
    r(\eta):=g(z+\eta)-g(z).    
\end{equation}
For 
\begin{equation}
    F\left(x,r(\eta),\theta\right)= dist (p(y|x,\theta)\|p(y|x+r(\eta),\theta)),
\end{equation}
we have
\begin{equation}
    \begin{aligned}
        \down^2_\eta F(x,r(\eta),\theta) &= (J_{z+\eta}g)^T \down^2_r F(x,r(\eta),\theta) (J_{z+\eta}g) \\
        + & \down^2_\eta  g(z+\eta) \cdot \down_r F(x,r(\eta),\theta).
    \end{aligned}
\end{equation}
While on the other hand, since $dist(\cdot,\cdot)$ is some distance measure with minimum zero and $r(0)=0$ is the corresponding optimal value, we have
\begin{equation}
    F(x,r(0),\theta) =0, \quad
    \down_r F(x,r(0),\theta) = 0.
\end{equation}
Therefore,
\begin{equation}
    \down^2_\eta F(x,r(0),\theta) = (J_z g)^T \down^2_r F(x,r(0),\theta) J_z g = J^T H J.
\end{equation}
Thus the targeted matrix vector product could be efficiently computed as 
\begin{equation}
    J^T H J \eta = \down^2_\eta F(x,r(0),\theta) \cdot \eta = \down_\eta \left(\down_\eta F(x,r(0),\theta)\cdot \eta\right).
\end{equation}
Note that $\down_\eta F(x,r(0),\theta)\cdot \eta$ is a scalar, hence the gradient of which could be obtained by back propagating the network for once. And it only costs twice back propagating for the computation of $J^T H J \eta$.

\subparagraph{Solving $J^T J\mu=v$.}
Similarly, define
\begin{equation}
    K(\eta):= \left(g(z+\eta)-g(z)\right)^T \left(g(z+\eta)-g(z)\right).
\end{equation}
We have
\begin{equation}
    \down^2_\eta K(\eta) = (J_{z+\eta}g)^T J_{z+\eta}g + \down^2_\eta g(z+\eta)\cdot K(\eta).
\end{equation}
Since $K(0)=0$, we have 
\begin{equation}
    \down^2_\eta K(0) = (J_z g)^T J_z g = J^T J.
\end{equation}
Thus the matrix vector product $J^T J \mu$ could be evaluated similarly as 
\begin{equation}
    J^T J \mu = \down_\eta \left(\down_\eta K(0) \cdot \mu\right).
\end{equation}
The extra cost for evaluating $J^T J \mu$ is still back propagating the network for twice. Due to $J^T J$ being positive definite ($g$ is non-degenerated), we can apply several steps of conjugate gradient to solve $J^T J \mu = v$ efficiently.

\par
By iterating Eq.~(\ref{eq:power-cg}), we obtain the optimal solution $\eta_{\parallel}$ of Eq.~(\ref{eq:tangent-vat-eigen}). The desired optimal solution is then $r_{\parallel} = \epsilon J\eta_{\parallel}/\|J\eta_{\parallel}\|$, hence $\Rcal_{\text{tangent}}(x;\theta) = F(x,r_{\parallel},\theta)$, which could be optimized by popular gradient optimizers.

%Compared with manifold regularization based on tangent propagation~\cite{simard1998transformation,kumar2017semi} or manifold Laplacian norm~\cite{belkin2006manifold,lecouat2018semi}, which is computationally inefficient due to the evaluation of Jacobian, our proposed TAR could be efficiently implemented, thanks to the low computational cost of virtual adversarial training.

\subsection{Normal adversarial regularization}
Motivated by the noisy observation assumption indicating that the observed data contains noise driving them off the underlying manifold, we further come up with the normal adversarial regularization (NAR) to enforce the robustness of the classifier against such noise, by performing virtual adversarial training in the normal space. The mathematical description is
\begin{equation}
    \begin{aligned}
        \Rcal_{\text{normal}} (x;\theta) :=& \max_{\norm{r}_2 \le \epsilon, r\bot T_x \Mcal} F(x,r,\theta) \\
        \approx & \max_{\norm{r}_2 \le \epsilon, r\bot T_x \Mcal} \half r^T H r.
    \end{aligned}
    \label{eq:ambient-vat}
\end{equation}

Note that $T_x \Mcal$ is spanned by the columns of $J=J_z g$, thus $r\bot T_x\Mcal \Leftrightarrow J^T\cdot r=0$. Therefore we could reformulate Eq.~(\ref{eq:ambient-vat}) as
\begin{equation}
    \begin{aligned}
        \underset{r\in\Rbb^D}{\text{maximize}} &\quad \half r^T H r,\\
        \text{s.t.} &\quad  \norm{r}_2\leq \epsilon,\quad J^T \cdot r = 0.
    \end{aligned}
    \label{eq:ambient-vat-opti}
\end{equation}
However, Eq.~(\ref{eq:ambient-vat-opti}) is not easy to optimize since $J^T\cdot r$ cannot be efficiently computed. To overcome this, instead of requiring $r$ being orthogonal to the whole tangent space $T_x \Mcal$, we take a step back to demand $r$ being orthogonal to only one specific tangent direction, i.e., the tangent space adversarial perturbation $r_{\parallel}$. Thus the constraint $J^T\cdot r=0$ is relaxed to $(r_{\parallel})^T\cdot r=0$.
% And Eq.~(\ref{eq:ambient-vat-opti}) becomes
% \begin{equation}
%     \begin{aligned}
%         \underset{r\in\Rbb^D}{\text{maximize}}\qquad & \half r^T H r \\
%         \text{s.t.}\qquad & \norm{r}_2\leq \epsilon,\quad  r_{\parallel}^T \cdot r = 0.
%     \label{eq:ambient-vat-relax}
%     \end{aligned}
% \end{equation}
And we further replace the constraint by a regularization term,
\begin{equation}
    \begin{aligned}
        \underset{r\in\Rbb^D}{\text{maximize}} &\quad  \half r^T H r - \lambda r^T (r_{\parallel} r_{\parallel}^T) r,\\
        \text{s.t.} &\quad  \norm{r}_2\leq \epsilon,
    \end{aligned}
    \label{eq:ambient-vat-approx}
\end{equation}
where $\lambda$ is a hyperparameter introduced to control the orthogonality of $r$.

Since Eq.~(\ref{eq:ambient-vat-approx}) is again an eigenvalue problem, and we can apply power iteration to solve it.
Note that a small identity matrix $\lambda \|r_{\parallel}\| I$ is needed to be added to keep $\half H-\lambda r_{\parallel} r_{\parallel}^T + \lambda \|r_{\parallel}\|I$ semi-positive definite, which does not change the optimal solution of the eigenvalue problem.
The power iteration is as
\begin{equation}
    r \leftarrow \half Hr - \lambda (r_{\parallel})^T r_{\parallel} r + \lambda \|r_{\parallel}\| r.
\end{equation}
And the evaluation of $Hr$ is by 
\begin{equation}
    H r = \down_r \left( \down_r F(x,0,\theta) \cdot r\right),
\end{equation}
which could be computed efficiently.
After finding the optimal solution of Eq.~(\ref{eq:ambient-vat-approx}) as $r_{\bot}$, the NAR becomes $\Rcal_{\text{normal}}(x,\theta) = F(x,r_{\bot},\theta)$.

\par
Finally, as suggested in~\cite{miyato2017virtual}, we add \emph{entropy regularization} to our loss function. It ensures neural networks to output more determinate predictions and has implicit benefits for performing virtual adversarial training.
\begin{equation}
    \Rcal_{\text{entropy}}(x,\theta):= -\sum_y p(y|x,\theta) \log p(y|x, \theta).
\end{equation}
Our final loss for SSL is
\begin{equation}
    \begin{aligned}
        L(D_l, D_{ul}, \theta) :=& \Ebb_{(x_l, y_l)\in\Dcal_{l}} \ell\left(y_l, p(y|x_l, \theta)\right) \\
        &+ \alpha_1 \Ebb_{x\in\Dcal} \Rcal_{\text{tangent}}(x,\theta)\\
        &+ \alpha_2 \Ebb_{x\in\Dcal} \Rcal_{\text{normal}}(x,\theta)\\
        &+ \alpha_3 \Ebb_{x\in\Dcal} \Rcal_{\text{entropy}}(x,\theta).
    \end{aligned}
\end{equation}

%The TAR inherits the computational efficiency from VAT and the manifold invariance property from traditional manifold regularization. The NAR causes the classifier for SSL being robust against the off manifold noise contained in the observed data. These advantages make our proposed TNAR, the combination of TAR and NAR, a reasonable regularization method for SSL, the superiority of which will be shown in the experiment part in Section~\ref{sec:exp}.

\section{Comparison to other methods}
\label{sec:compare}
\begin{figure}
    \centering
    \includegraphics[width=0.49\linewidth]{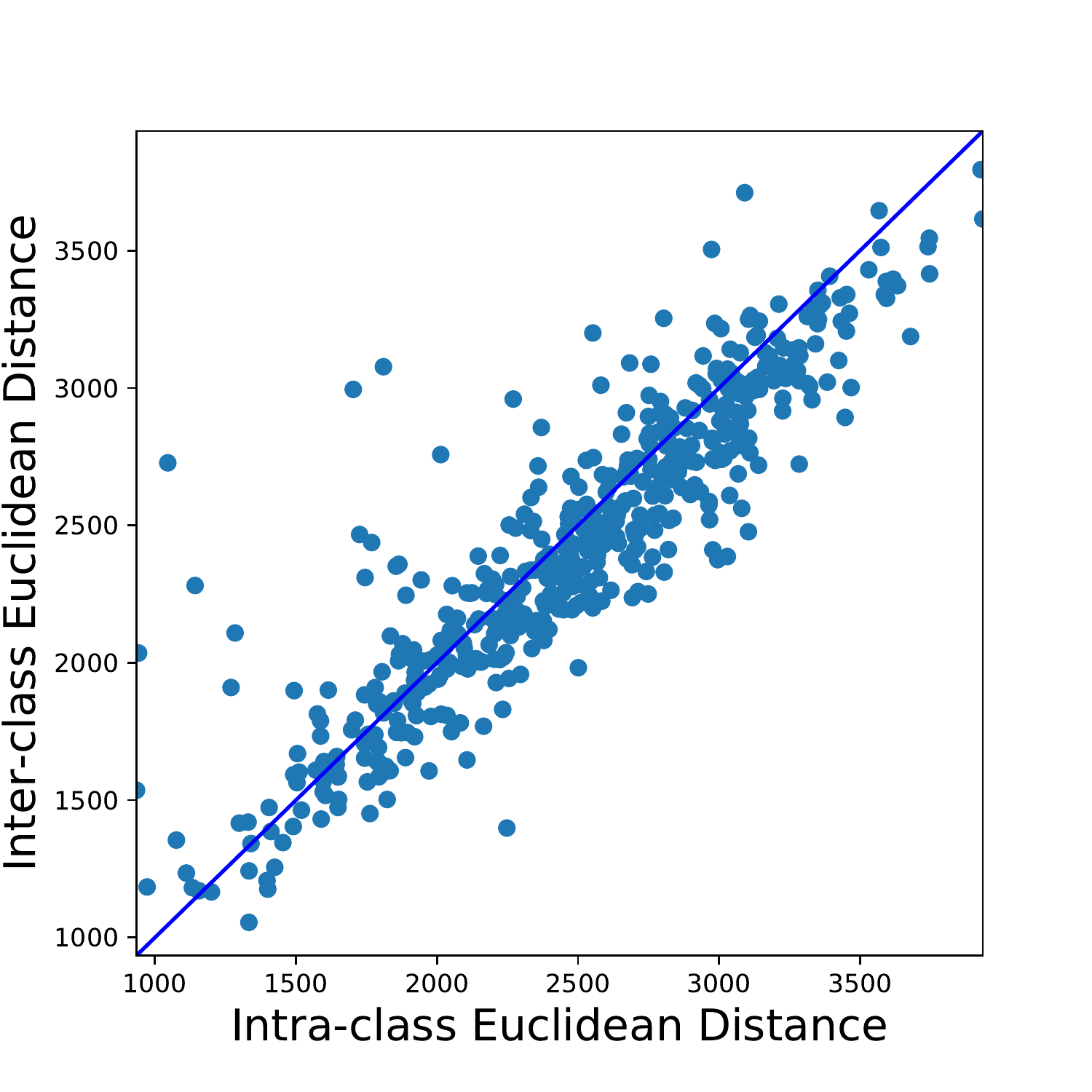}
    \includegraphics[width=0.49\linewidth]{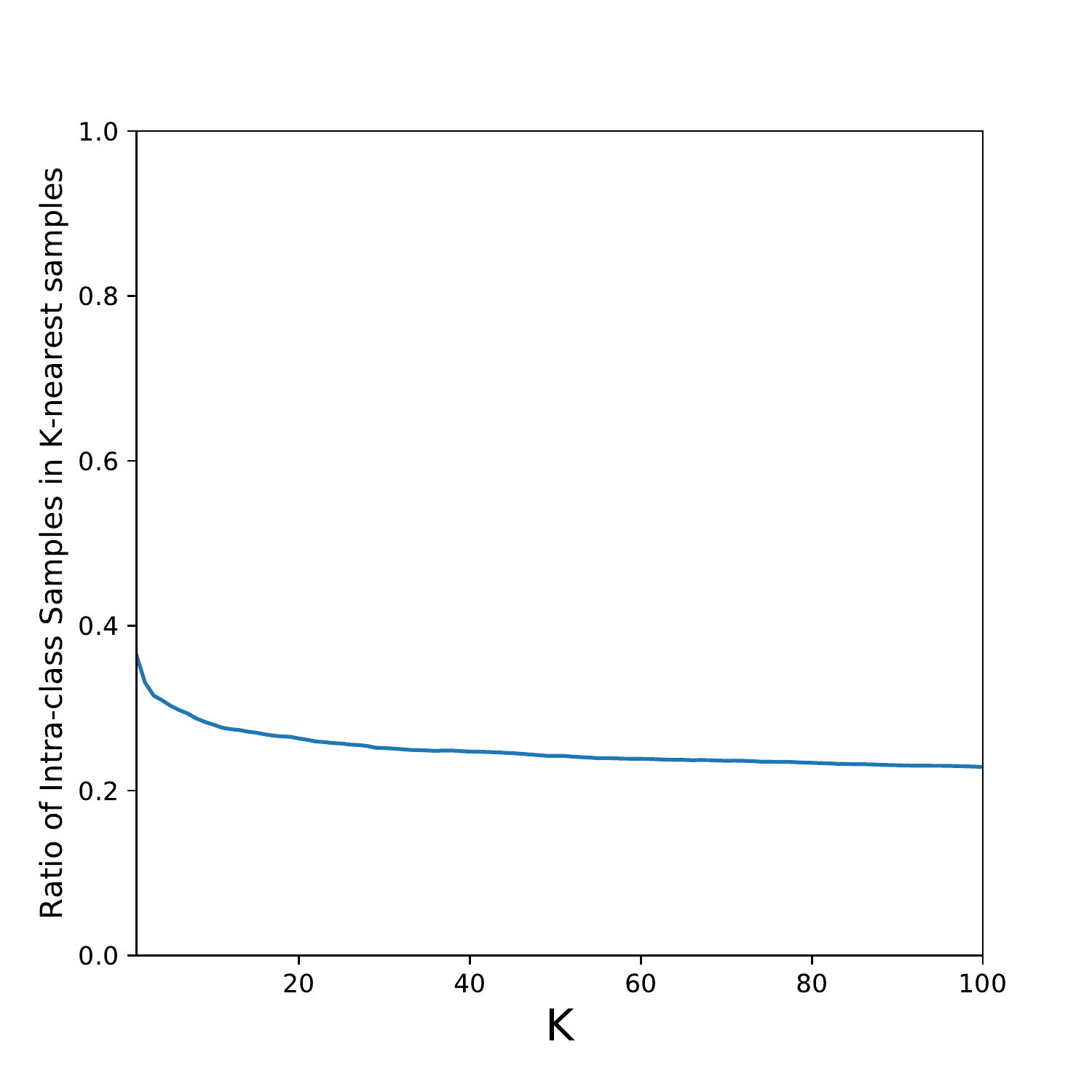}
    \caption{Left: the smallest intra-class distance vs. the smallest inter-class distance for CIFAR-10 dataset.
    X-axis: the smallest Euclidean distance to the other examples of the same class.
    Y-axis: the smallest Euclidean distance to the other examples of the different class.
    We only plot such coordinate for $500$ examples.
    Right: the ratio of intra-class example among its $K$-nearest neighborhood for CIFAR-10 dataset. The ratio is averaged over $500$ examples.
    From the figures we clearly see that for most examples, 1) the smallest inter-class distance is shorter or at least about the same scale as the least intra-class distance, and 2) its $K$-nearest neighborhood contains more inter-class examples than intra-class examples. 
    }
    \label{fig:cifar-distance}
\end{figure}

\paragraph{Virtual adversarial training}
Our proposed TNAR serves as an extension of VAT, by taking the information of data manifold into consideration.
% In terms of regularization effect,
VAT equally penalizes the smoothness along each dimension of  the whole observation space, not discriminating different directions.
In contrast, TNAR enforces the smoothness of the classifier along the manifold and orthogonal to the manifold separately.
This separate treatment along the two directions allows TNAR to impose different scales of smoothness along the tangent space and the normal space of the data manifold,
which is particularly crucial for inducing desired regularization effect.
To illustrate this, considering an image sample, its Euclidean neighborhood in the input space could contain many inter-class samples, besides intra-class ones, as demonstrated in Figure~\ref{fig:cifar-distance}.
Thus the output of the ideal classifier must vary significantly inside such Euclidean neighborhood to correctly classify the contained samples, 
which makes it essentially improper for VAT to enforce that the classifier does not change much inside this Euclidean ball. 
A more reasonable treatment is to adopt the manifold assumption and impose different scales of smoothness of the classifier along the manifold and its orthogonal direction, as TNAR has done.

% Similar to adversarial training with FGSM~\cite{goodfellow2014explaining}, the computational complexity of VAT is also cheap. The number of extra times of back propagation for VAT depends on the number of power iteration, and typically once is enough~\cite{miyato2016adversarial} (corresponding to back propagation twice). The crucial benefit is that the computational cost is independent of the dimensionality of the input or output of the classifier, making it particularly suitable for modern high-dimensional datasets.

% In terms of computational complexity,
% TNAR adds constant times of back propagation each iteration to VAT. In our implementation, the TAR term requires $10$ times of back propagation (corresponding to four times conjugate gradient iteration and once VAT iteration), and the NAR terms requires $2$ times of back propagation (corresponding to once VAT iteration).
% Thus TNAR adds $10$ more back propagation to VAT, which adds $2$ times of back propagation compared to supervised learning objectives.
% The important thing is, the extra costs of both TNAR and VAT are constant to the dimensionality of data and the number of classes, thus suitable to modern complex semi-supervised learning problems.

\paragraph{Jacobian based manifold regularization}
As explained in Eq.~(\ref{eq:tangent-prop}) and Eq.~(\ref{eq:laplacian}), tangent propagation~\cite{simard1998transformation,kumar2017semi} and manifold Laplacian norm~\cite{belkin2006manifold,lecouat2018semi,qi2018global} are also popular methods for realizing manifold regularization for SSL.
However, our TNAR is the first to use VAT constructing manifold regularization.
The difference between TNAR and Jacobian norm based manifold regularization is two folds.

Firstly, they lead to different manifold smoothness conditions on the classifier.
Tangent propagation and manifold Laplacian norm smooth the classifier by regularizing its norm of the manifold Jacobian.
TNAR, on the other hand, smooths the classifier through penalizing the virtual adversarial loss defined by the distance of an example with its tangent directional virtual adversarial example. This involves the second order information of the virtual adversarial loss along the manifold. 
Theoretically, it is not easy to say that one smoothness is superior to the other.
Nonetheless, empirical experiments on multiple datasets (Section~\ref{sec:exp}) suggest that our proposed TNAR achieves better performance on SSL.
We leave the theoretical analysis as future work.

Secondly, as shown in Eq.~(\ref{eq:tangent-prop}) and Eq.~(\ref{eq:laplacian}),
all the existing Jacobian based manifold regularization requires evaluating the Jacobian of either the classifier, or the generator as manifold coordinate chart,
which is prohibitively feasible for modern high-dimensional datasets given large neural networks.
Alternatively, some works suggested stochastically evaluating these Jacobian based regularization terms.
Kumar at.el.~\cite{kumar2017semi} proposed to randomly preserve several columns of $J_z g$ as the approximation of the tangent space $T_x \Mcal$, and Lecouat at.el.~\cite{lecouat2018semi} applied the norm of several directional gradients to approximate the norm of the Jacobian.
However, such stochastic strategies, unfortunately with high variance, could cause implicit side affects on the manifold smoothness of the classifier.
Compared with them, the computational cost of our proposed TNAR does not rely on the dimensionality of the datasets,
since performing VAT only requires several times of power iteration (typically once), and TNAR adds constant extra times of back or forward propagation to VAT.
This advantage makes TNAR a potentially better manifold regularization method for modern semi-supervised learning tasks.

% The main issue of the Jacobian based regularizer is the computational cost.
% Suppose the output of classifier $f$ is $K$-dimensional, the observed data $x \in \Rbb^D$ and the hidden represent $z\in R^d, d\ll D$, 
% then $J_x f(x)\in \Rbb^{K\times D}, J_z f(g(z)) \in \Rbb^{K\times d}$, and the tangent space $T_x M\cong \Rbb^{d}$ since it is the span of the columns of the Jacobian $J_z g(z)$.
% Therefore,  evaluating Eq.~(\ref{eq:tangent-prop}) requires at least $K$ times back propagation of network $f$
% \footnote{In popular deep learning frameworks like tensorflow or pytorch, it is only supported to directly compute the gradient of scalar function. Thus to evaluate the Jacobian of $\phi: \Rbb^m \to \Rbb^n$, one needs $n$ times back propagating to compute the gradient of each component of $\phi$ respectively. Or one could estimate the Jacobian by (central) differences, e.g., $\frac{\partial \phi(x)}{\partial r} = \frac{\partial \phi(x+r)-\partial \phi(x-r)}{2r}$, which requires $2m$ times of forward propagation to estimate $m$ directional gradients. In the time cost analysis of this work, we adopt the first method. However, the conclusion holds same using the second approach.}
% and $D$ times back propagation of network $g$, and  evaluating Eq.~(\ref{eq:laplacian}) costs at least $K$ times back propagation of both network $f$ and $g$.

\paragraph{Other approaches for SSL}
There is also a wide class of SSL framework based on GAN~\cite{salimans2016improved,odena2016semi,dumoulin2016adversarially,dai2017good}.
Most of them modify the discriminator to include a classifier, by splitting the real class of original discriminator into $K$ subclasses, where $K$ is the number of classes of labeled data.
The features extracted for distinguishing the example being real or fake, which can be viewed as a kind of coarse label, have implicit benefits for supervised classification task.
Though in TNAR, GAN with encoder could be adopted as a method to identify the underlying manifold, these two kinds of approaches are motivated from different perspectives.
TNAR focuses on the manifold regularization other than the feature sharing as in the GAN frameworks for SSL.

Besides above, there are also other strategies for SSL, e.g., Tripple GAN~\cite{li2017triple}, Mean Teacher~\cite{tarvainen2017mean}, $\Pi$ model~\cite{laine2016temporal}, CCLP~\cite{kamnitsas2018semi} etc.
We leave the comparison of the performance with TNAR in Section~\ref{sec:exp}.

\section{Experiments}
\label{sec:exp}
To demonstrate the advantages of our proposed TNAR for SSL, we conduct a series of experiments on both artificial and real datasets.
The tested TNAR based methods for SSL include:
\begin{description}
    \item[TNAR-VAE]: TNAR with the underlying manifold estimated by VAE;
    \item[TNAR-LGAN]: TNAR with the underlying manifold estimated by localized GAN;
    \item[TNAR-Manifold]: TNAR with oracle underlying manifold for the observed data, only used for artificial dataset;
    \item[TNAR-AE]: TNAR with the underlying manifold estimated roughly by autoendoer, only used for artificial dataset.
    \item[TAR]: tangent adversarial regularization for ablation study.
    \item[NAR]: normal adversarial regularization for ablation study.
\end{description}
If not stated otherwise, all the above methods contain entropy regularization term.
% 1) {TNAR-VAE}: the proposed TNAR method, with the underlying manifold estimated by VAE;
% 2) {TNAR-LGAN}: the proposed TNAR method, with the underlying manifold estimated by localized GAN;
% 3) {TNAR-Manifold}: the proposed TNAR method with oracle underlying manifold for the observed data, only used for artificial dataset;
% 4) {TNAR-AE}: the proposed TNAR method, with the underlying manifold estimated roughly by autoendoer, only used for artificial dataset.
% 5) {TAR}: the tangent adversarial regularization for ablation study.
% 6) {NAR}: the normal adversarial regularization for ablation study.
% If not stated otherwise, all the above methods contain entropy regularization term.

\subsection{Two-rings artificial dataset}
We first introduce experiments on a two-rings artificial dataset to show the effectiveness of TNAR intuitively.
In this experiments, there is $3,000$ unlabeled data (gray dots) and $6$ labeled data (blue dots), $3$ for each class. The detailed construction could be found in Supplementary Materials.

\begin{figure}
    \centering
    \includegraphics[width=0.6\linewidth]{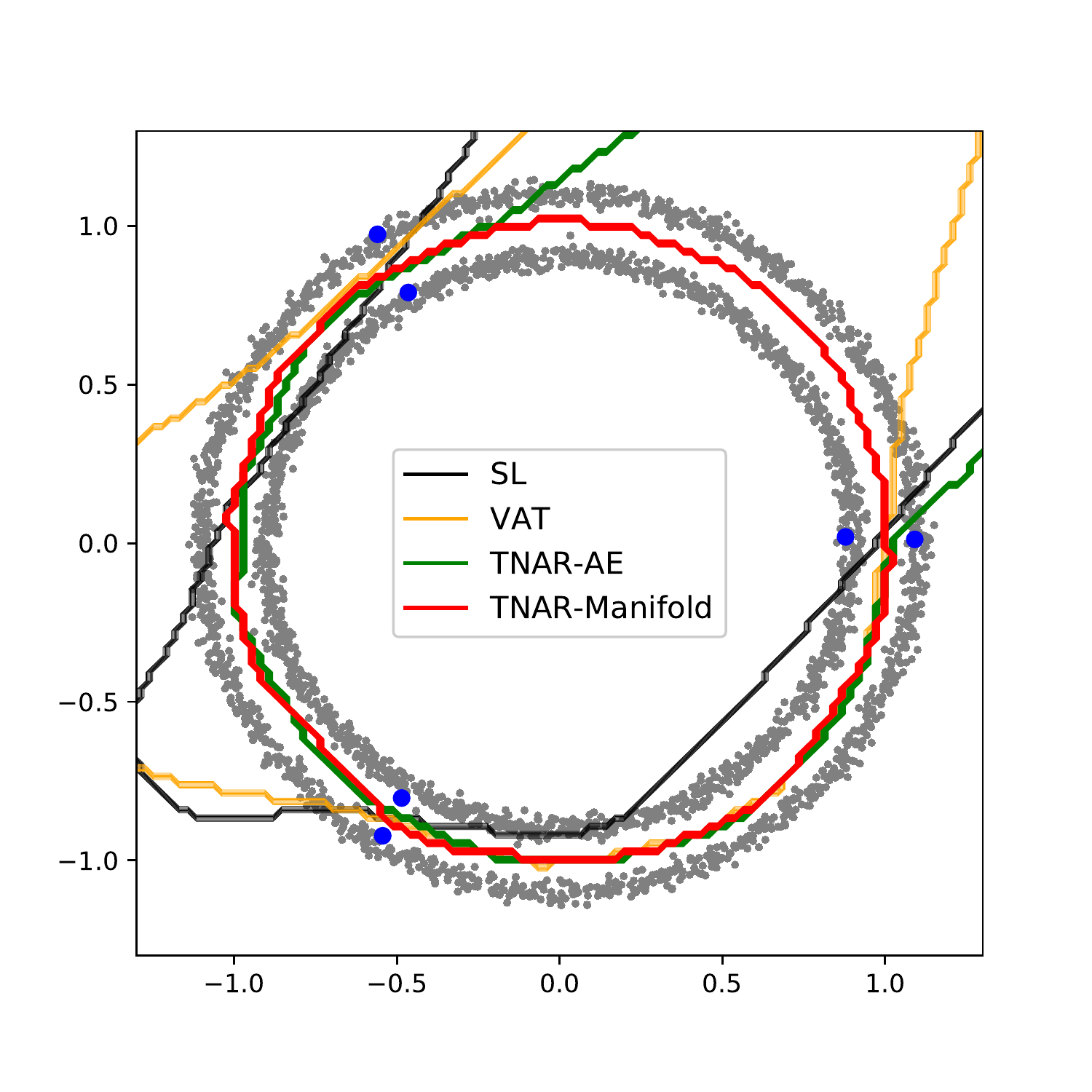}
    \caption{The decision boundaries of compared methods on two-rings artificial dataset.
    Gray dots distributed on two rings: the unlabeled data.
    Blue dots ($3$ in each ring): the labeled data.
    Colored curves: the decision boundaries found by compared methods.}
    \label{fig:2rings}
\end{figure}

The performance of each compared methods is shown in Table~\ref{tb:2rings}, and the corresponding classification boundary is demonstrated in Figure~\ref{fig:2rings}. The TNAR under true underlying manifold (TNAR-Manifold) perfectly classifies the two-rings dataset with merely $6$ labeled data, while the other methods fail to predict the correct decision boundary.
The failure of VAT supports our claims of its shortcut in Section~\ref{sec:compare}.
Even with the underlying manifold roughly approximated by an autoendoer, our approach (TNAR-AE) outperforms VAT in this artificial dataset. However, the performance of TNAR-AE is worse than TNAR-Manifold, indicating that the effectiveness of TNAR relies on the quality of estimating the underlying manifold.

\begin{table}
    \centering
    \caption{Classification errors ($\%$) of compared methods on two-ring artificial dataset.
    We test with and without entropy regularization in each method and report the best one. In VAT and TNAR-AE, without entropy regularization is better; For TNAR-Manifold, adding entropy regularization is better.}
    \label{tb:2rings}
    \begin{tabular}{ll}
        \hhline{==}
        Model & Error (\%)\\
        \hline
        Labeled data only & 32.95 \\
        \hline
        VAT & 23.80 \\
        \hline
        % VAT (ent) & 42.00 \\
        % \hline
        TNAR-AE  & 12.45 \\
        \hline
        % TNAR-AE (ent)  & 38.60 \\
        % \hline
        TNAR-Manifold & 9.90 \\
        \hline
        TNAR-Manifold (ent) & 0 \\
        \hhline{==}
    \end{tabular}   
\end{table}

\subsection{FashionMNIST}
We also conduct experiments on FashionMNIST dataset\footnote{\url{https://github.com/zalandoresearch/fashion-mnist}}.
There are three sets of experiments with the number of labeled data being $100$, $200$ and $1,000$, respectively. The details about the networks are in Supplementary Materials.
\begin{table*}
    \small
    \caption{Classification errors ($\%$) of compared methods on FashionMNIST dataset.}     \label{tb:fashion}
    \centering
    \begin{tabular}{cccc} \hhline{====}
        Method & 100 labels & 200 labels & 1000 labels \\
        \hline
        VAT & 27.69 & 20.85 & 14.51 \\
        \hline
        TNAR/TAR/NAR-LGAN & 23.65/24.87/28.73 & 18.32/19.16/24.49 & 13.52/14.09/15.94 \\
        \hline
        TNAR/TAR/NAR-VAE & \textbf{23.35}/26.45/27.83 & \textbf{17.23}/20.53/24.81 & \textbf{12.86}/14.02/15.44 \\
        \hhline{====}
    \end{tabular}
\end{table*}

The corresponding results are shown in Table~\ref{tb:fashion}, from which we observe at least two phenomena.
The first is that our proposed TNAR methods (TNAR-VAE, TNAR-LGAN) achieve lower classification errors than VAT in all circumstances with different number of labeled data.
The second is that the performance of our method depends on the estimation of the underlying manifold of the observed data. In this case, TNAR-VAE brings larger improvement than TNAR-LGAN, since VAE produces better diverse examples according to our observation.
% As the development of generative model capturing more accurate underlying manifold, it is expected that our proposed regularization strategy benefits more for SSL.

\subsection{CIFAR-10 and SVHN}
There are two classes of experiments for demonstrating the effectiveness of TNAR in SSL, SVHN with $1,000$ labeled data, and CIFAR-10 with $4,000$ labeled data. The experiment setups are identical with \cite{miyato2017virtual}. We test two kinds of convolutional neural networks as classifier (denoted as "small" and "large") as in~\cite{miyato2017virtual}.
We test both VAE and Localized GAN as the underlying data manifold. More detailed experimental settings are included in Supplementary Materials.
We test the performance of TNAR with or without data augmentation, with the identical augmentation strategy used in~\cite{miyato2017virtual}.
Note that when perform TNAR with data augmentation, the corresponding data manifold should also be trained with data augmentation.
It is worth to remark that VAT~\cite{miyato2017virtual} and VAT + SNTG~\cite{luo2017smooth} adopts ZCA as pre-processing on CIFAR-10 experiments, while we do not use this trick implementing TNAR experiments.

\begin{table}
    \caption{Classification errors ($\%$) of compared methods on SVHN and CIFAR-10 datasets without data augmentation.}
    \label{tb:cifar}
    \centering
    \small
    \begin{tabular}{p{3.6cm}p{1.6cm}p{1.6cm}}
        \hhline{===}
        Method & SVHN 1,000 labels & CIFAR-10 4,000 labels \\
        \hline
        VAT (small)~\cite{miyato2017virtual} & $6.83\pm 0.24$ & $14.87\pm 0.13$ \\
        VAT (large)~\cite{miyato2017virtual} & $4.28\pm 0.10$ & $13.15\pm 0.21$ \\
        VAT + SNTG~\cite{luo2017smooth} & $4.02\pm 0.20$ & $12.49\pm 0.36$ \\
        $\Pi$ model~\cite{laine2016temporal} & $5.43\pm 0.25$ & $16.55\pm 0.29$ \\
        Mean Teacher~\cite{tarvainen2017mean} & $5.21\pm 0.21$ & $17.74\pm 0.30$ \\
        CCLP~\cite{kamnitsas2018semi} & $5.69\pm 0.28$ & $18.57\pm 0.41$ \\
        ALI~\cite{dumoulin2016adversarially} & $7.41\pm 0.65$ & $17.99\pm 1.62$ \\
        Improved GAN~\cite{salimans2016improved} & $8.11\pm 1.3$ & $18.63\pm 2.32$ \\
        Tripple GAN~\cite{li2017triple} & $5.77\pm 0.17$ & $16.99\pm 0.36$ \\
        Bad GAN~\cite{dai2017good} & $4.25\pm 0.03$ & $14.41\pm 0.30$ \\
        LGAN~\cite{qi2018global} & $4.73\pm 0.16$ & $14.23\pm 0.27$ \\
        \footnotesize{Improved GAN + JacobRegu + tangent}~\cite{kumar2017semi} & $4.39\pm 1.20$ & $16.20\pm 1.60$ \\
        \footnotesize{Improved GAN + ManiReg}~\cite{lecouat2018semi} & $4.51\pm 0.22$ & $14.45\pm 0.21$ \\
        \hline
        TNAR-LGAN (small) & $4.25\pm 0.09$ & $12.97\pm 0.31$ \\
        TNAR-LGAN (large) & $4.03\pm 0.13$ & $12.76\pm 0.04$ \\
        TNAR-VAE (small) & $3.99\pm 0.08$ & $12.39\pm 0.11$ \\
        TNAR-VAE (large) & $\mathbf{3.80\pm 0.12}$ & $\mathbf{12.06\pm 0.35}$ \\
        TAR-VAE (large) & $5.62\pm 0.19$ & $13.87\pm 0.32$ \\
        NAR-VAE (large) & $4.05\pm 0.04$ & $15.91\pm 0.09$ \\
        \hhline{===}
    \end{tabular}
\end{table}

In Table~\ref{tb:cifar} we report the experiments results on SVHN and CIFAR-10, without data augmentation.
And in Table~\ref{tb:cifar-aug} the results on SVHN and CIFAR-10 with data augmentation are presented.
The comparison demonstrates that our proposed TNAR outperforms all the other state-of-the-art SSL methods as far as we known on both SVHN and CIFAR-10, with or without data augmentation.
Especially, compared with VAT or manifold regularization like Improved GAN + JacobRegu + tagent~\cite{kumar2017semi} or Improved GAN + ManiReg~\cite{lecouat2018semi},
TNAR brings an evident improvements to them, as our analysis in Section~\ref{sec:compare} has suggested.
Similar to experiments on FashionMNIST datasets, we observe that for TNAR, the underlying manifold identified by VAE benefits more than the manifold identified by Localized GAN. We attribute this phenomenon to the relatively lacking of diversity of the images generated by Localized GAN.

\begin{table}
    \caption{Classification errors ($\%$) of compared methods on SVHN and CIFAR-10 datasets with data augmentation.}
    \label{tb:cifar-aug}
    \centering
    \small
    \begin{tabular}{lp{1.6cm}p{1.6cm}}
        \hhline{===}
        Method & SVHN 1,000 labels & CIFAR-10 4,000 labels \\
        \hline
        VAT (large)~\cite{miyato2017virtual} & $3.86\pm 0.11$ & $10.55\pm 0.05$ \\
        VAT + SNTG~\cite{luo2017smooth} & $3.83\pm 0.22$ & $9.89\pm 0.34$ \\
        $\Pi$ model~\cite{laine2016temporal} & $4.82\pm 0.17$ & $12.36\pm 0.31$ \\
        Temporal ensembling~\cite{laine2016temporal} & $4.42\pm 0.16$ & $12.16\pm 0.24$ \\
        Mean Teacher~\cite{tarvainen2017mean} & $3.95\pm 0.19$ & $12.31\pm 0.28$ \\
        LGAN~\cite{qi2018global} & - & $9.77\pm 0.13$ \\
        \hline
        TNAR-VAE (large) & $\mathbf{3.74\pm 0.04}$ & $\mathbf{8.85\pm 0.03}$ \\
        \hhline{===}
    \end{tabular}
\end{table}

\subsection{Ablation study}
We conduct ablation study on FashionMNIST, SVHN and CIFAR-10 datasets to demonstrate that both of the two regularization terms in TNAR are crucial for SSL.
The results are reported in Table~\ref{tb:fashion} and the last two lines in Table~\ref{tb:cifar}.
Removing either tangent adversarial regularization or normal adversarial regularization will harm the final performance, since they fail to enforce the manifold invariance or the robustness against the off-manifold noise.
In together, the proposed TNAR achieves the best performance.

\begin{figure}
    \centering
    \begin{tabular}{ccccc}
        \includegraphics[width=0.14\linewidth]{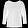}
        & \includegraphics[width=0.14\linewidth]{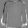}
        & \includegraphics[width=0.14\linewidth]{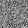}
        & \includegraphics[width=0.14\linewidth]{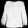}
        & \includegraphics[width=0.14\linewidth]{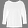}
        \\ \includegraphics[width=0.14\linewidth]{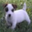}
        & \includegraphics[width=0.14\linewidth]{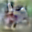}
        & \includegraphics[width=0.14\linewidth]{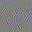}
        & \includegraphics[width=0.14\linewidth]{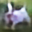}
        & \includegraphics[width=0.14\linewidth]{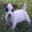}
        % \\ (a) & (b) & (c) & (d) & (e)
    \end{tabular}
    \caption{The perturbations and adversarial examples in the tangent space and the normal space.
        Note that the perturbations is actually too small to distinguish easily, thus we show the scaled perturbations.
    First row: FashionMNIST dataset; Second row: CIFAR-10 dataset.
    From left to right: original example, tangent adversarial perturbation, normal adversarial perturbation,
    tangent adversarial example, normal adversarial example.}
    \label{fig:adversarial}
\end{figure}

Furthermore, the adversarial perturbations and adversarial examples from FashionMNIST and CIFAR-10 are shown in Figure~\ref{fig:adversarial}.
We can easily observe that the tangent adversarial perturbation focuses on the edges of foreground objects, while the normal space perturbation mostly appears as certain noise over the whole image.
This is consistent with our understanding on the role of perturbation along the two directions that capture the different aspects of smoothness.

\section{Discussion}
\label{sec:discuss}
As shown in our experiments, the data manifold is crucial for the improvement of our proposed TNAR.
Though TNAR seems to work with a wide range of manifold coordinate chart, e.g., VAE and Localized GAN, it is still not clear which kind of manifold benefits most for TNAR.
Dai at.el~\cite{dai2017good} suggested that a bad generator works better for GAN based framework for semi-supervised learning.
Our experiments agree with this argument to some extent. Localized GAN could produce detailed images than VAE, but the latter cooperates better with TNAR in all our experiments.
At current stage, we conjecture that a more diverse generator helps more for TNAR, since diversity on generator enables TNAR to explore more different directions along the data manifold.
The throughout analysis is left for further work.

\section{Conclusion}
\label{sec:conclude}
We present the tangent-normal adversarial regularization for semi-supervised learning, a novel regularization strategy based on virtual adversarial training and manifold regularization.
TNAR is composed of regularization on the tangent and normal space separately.
The tangent adversarial regularization enforces manifold invariance of the classifier,
while the normal adversarial regularization imposes robustness of the classifier against the noise contained in the observed data.
Experiments on artificial dataset and multiple practical datasets demonstrate that our approach outperforms other state-of-the-art methods for semi-supervised learning.
% DO NOT say anything about the benefits of the breakthrough of generative models ....
% since it benefits MANIFOLD REGULARIZATION, but not only to us.

{\small
\bibliographystyle{ieee}
% \bibliography{egbib}
\bibliography{bibliography}
}

\appendix
\section{Two-rings dataset}
The underlying manifold for two-rings data is given by $\Mcal=\Mcal_+ \cup \Mcal_-$, where
\begin{align*}
    \Mcal_+ &= \left\{(x_1,x_2)\left|x_1^2+x_2^2=0.9^2\right.\right\} \\
    \Mcal_- &= \left\{(x_1,x_2)\left|x_1^2+x_2^2=1.1^2\right.\right\} .
\end{align*}
The observed data is sampled as $x = x_0 + n$, where $x_0$ is uniformly sampled from $\Mcal$ and $n \sim \Ncal (0, 2^{-2})$.
We sample $6$ labeled training data, $3$ for each class, and $3,000$ unlabeled training data, as shown in Figure~3.

\section{Experiments details on FashionMNIST}
In FashionMNIST\footnote{\url{https://github.com/zalandoresearch/fashion-mnist}} experiments, we preserve $100$ data points for validation from the original training dataset. That is, we use $100/200/1,000$ labeled data for training and another $100$ labeled data for validation.
For pre-processing, we scale pixel values of images into $[0,1]$.
The architecture of the neural network for classification  is as following: $(a,b)$ denotes the convolution filter is with $a\times a$ shape and $b$ channels. The max pooling layer is with stride $2$. And we apply local response normalization (LRN)~\cite{robinson2007explaining}. The number of hidden nodes in the first fully connected layer is $512$.
\begin{align*}
    &\text{Conv} (3,32) \to \text{ReLU} \to \text{Conv} (3,32) \to \text{ReLU} \to \\
    &\text{MaxPooling} \to \text{LRN} \to \text{Conv} (3,64) \to \text{ReLU} \to \\
    &\text{Conv} (3,64) \to \text{ReLU} \to \text{MaxPooling} \to \text{LRN}\\
    &\to \text{FC1} \to \text{ReLU} \to \text{FC2}
\end{align*}

For the labeled data, the batch size is $32$, and for the unlabeled data, the batch size is $128$.
All networks are trained for $12,000$ updates.
The optimizer is ADAM with initial learning rate $0.001$, and linearly decayed over the last $4,000$ updates.
The hyperparameters tuned is the magnitude of the tangent adversarial perturbation $\epsilon_1$, the magnitude of the normal adversarial perturbation $\epsilon_2$
and the hyperparameter $\lambda$ in Eq.~(28). All other hyperparameters are  set as $1.0$.
We tune $\lambda$ from $\{1,0.1,0.01,0.001\}$, and $\epsilon_1, \epsilon_2$ randomly from $[0.05, 20]$.
The corresponding hyperparameters used for experiments in the main paper are reported in Table~\ref{tb:fashion-hyper}.

\begin{table*}
    \small
    \caption{The hyperparameters for TNAR on FashionMNIST dataset.
    $\lambda$: the hyperparameter in Eq.~(28);$\epsilon_1$ and $\epsilon_2$: the norms of tangent adversarial perturbation and normal adversarial perturbation.}
    \label{tb:fashion-hyper}
    \centering
    \begin{tabular}{cccc} \hhline{====}
        Hyperparameters & 100 labels & 200 labels & 1000 labels \\
        \hline
        $\lambda/\epsilon_1/\epsilon_2$ for TNAR-LGAN & 0.1/0.2/8.0 & 0.1/4.0/0.5 & 1.0/20.0/0.5 \\
        \hline
        $\lambda/\epsilon_1/\epsilon_2$ for TNAR-VAE & 1.0/2.0/0.05 & 1.0/0.5/0.05 & 1.0/5.0/6.0 \\
        \hhline{====}
    \end{tabular}
\end{table*}

The encoder of the VAE for identifying the underlying manifold is a LeNet-like one, with two convolutional layers and one fully connected layer. And the decoder is symmetric with the encoder, except using deconvolutional layers to replace convolutional ones.
The latent dimensionality is $128$. 
The localized GAN for identifying the underlying manifold is similar as stated in~\cite{qi2018global}. And the implementation is modified from \url{https://github.com/z331565360/Localized-GAN}. We change the latent dimensionality into $128$.

VAE is pretrained and fixed during the training of TNAR. We tried both jointly and separately training the LGAN with the classifier, observing no significant difference.

\section{Experiments details on SVHN and CIFAR-10}
In SVHN\footnote{\url{http://ufldl.stanford.edu/housenumbers/}} and CIFAR-10\footnote{\url{https://www.cs.toronto.edu/~kriz/cifar.html}} experiments, we preserve $1,000$ data for validation from the original training set. That is, we use $1,000/4,000$ labeled data for training and another $1,000$ labeled data for validation.
The only pre-processing on data is to scale the pixels value into $[0,1]$.
We do not use data augmentation.
The structure of classification neural network is shown in Table~\ref{tab:cnn_models}, which is identical as in \cite{miyato2017virtual}.

For the labeled data, the batch size is $32$, and for the unlabeled data, the batch size is $128$.
For SVHN, all networks are trained for $48,000$ updates. And for CIFAR-10, all networks are trained for $200,000$ updates.
The optimizer is ADAM with initial learning rate $0.001$, and linearly decayed over the last $16,000$ updates.
The hyperparameters tuned is the magnitude of the tangent adversarial perturbation $\epsilon_1$, the magnitude of the normal adversarial perturbation $\epsilon_2$
and the hyperparameter $\lambda$ in Eq.~(28). Other hyperparameters are all set to $1$.
We tune $\lambda$ from $\{1,0.1,0.01,0.001\}$, and $\epsilon_1, \epsilon_2$ randomly from $[0.05, 20]$.

\begin{table*}
		\centering
		\caption{\label{tab:cnn_models} The structure of convolutional neural networks for experiments on CIFAR-10 and SVHN, based on \cite{springenberg2014striving,salimans2016improved, laine2016temporal}. 
        All the convolutional layers and fully connected layers are followed by batch normalization except the fully connected layer on CIFAR-10. The slopes of all lReLU functions in the networks are $0.1$.}
		\begin{tabular}{c|c|c}
			\toprule
			Conv-Small on SVHN & Conv-Small on CIFAR-10 & Conv-Large\\
			\midrule
		\multicolumn{3}{c}{32$\times$32 RGB image}  \\
		\midrule	
            3$\times$3 conv. 64 lReLU 		&3$\times$3 conv. 96 lReLU 		& 3$\times$3 conv. 128 lReLU 	\\
            3$\times$3 conv. 64 lReLU 		&3$\times$3 conv. 96 lReLU 		& 3$\times$3 conv. 128 lReLU 	\\
            3$\times$3 conv. 64 lReLU		&3$\times$3 conv. 96 lReLU		& 3$\times$3 conv. 128 lReLU	\\
		\midrule
		\multicolumn{3}{c}{2$\times2$ max-pool, stride 2}  \\
		\multicolumn{3}{c}{dropout, $p=0.5$}  \\
            \midrule   
            3$\times$3 conv. 128 lReLU 		&3$\times$3 conv. 192 lReLU 		& 3$\times$3 conv. 256 lReLU 	\\
            3$\times$3 conv. 128 lReLU 		&3$\times$3 conv. 192 lReLU 		& 3$\times$3 conv. 256 lReLU 	\\
            3$\times$3 conv. 128 lReLU 		&3$\times$3 conv. 192 lReLU 		& 3$\times$3 conv. 256 lReLU 	\\
            \midrule  
		\multicolumn{3}{c}{2$\times2$ max-pool, stride 2}  \\
		\multicolumn{3}{c}{dropout, $p=0.5$}  \\
            \midrule 
            3$\times$3 conv. 128 lReLU 		&3$\times$3 conv. 192 lReLU 		& 3$\times$3 conv. 512 lReLU 	\\
            1$\times$1 conv. 128 lReLU 		&1$\times$1 conv. 192 lReLU 		& 1$\times$1 conv. 256 lReLU 	\\
            1$\times$1 conv. 128 lReLU 		&1$\times$1 conv. 192 lReLU 		& 1$\times$1 conv. 128 lReLU 	\\
                \midrule
		\multicolumn{3}{c}{global average pool, 6$\times$6 $\rightarrow$ 1$\times$1 }  \\
            \midrule   
            dense 128 $\rightarrow$ 10	 		&dense 192$\rightarrow$ 10 	&dense 128$\rightarrow$ 10 \\
            \midrule
		\multicolumn{3}{c}{10-way softmax}  \\
		\bottomrule
		\end{tabular}
\end{table*}

The VAE for identifying the underlying manifold for SVHN and CIFAR-10 is implemented as in \url{https://github.com/axium/VAE-SVHN}. The only modification is we change the coefficient of the regularization term from $0.01$ to $1$.
The localized GAN for learning the underlying manifold for SVHN and CIFAR-10 is similar as stated in~\cite{qi2018global}. And the implementation is modified from \url{https://github.com/z331565360/Localized-GAN}.
We change the latent dimensionality of VAE and localized GAN into $512$ for both SVHN and CIFAR-10.
The hyperparameters used in the main paper for TNAR are reported in Table~\ref{tb:cifar-hyper}.

\begin{table*}
    \caption{The hyperparameters for TNAR on SVHN and CIFAR-10 datasets.
    $\lambda$: the hyperparameter in Eq.~(28);$\epsilon_1$ and $\epsilon_2$: the norms of tangent adversarial perturbation and normal adversarial perturbation.}
    \label{tb:cifar-hyper}
    \centering
    \small
    \begin{tabular}{p{5cm}p{2cm}p{2cm}p{3cm}p{3cm}}
        \hhline{=====}
        Hyperparameters & SVHN 1,000 labels & CIFAR-10 4,000 labels & SVHN 1,000 labels with augmentation & CIFAR-10 4,000 labels with augmentation\\
        \hline
        $\lambda/\epsilon_1/\epsilon_2$ for TNAR-LGAN (small) & 0.01/0.5/2.0 & 0.1/5.0/1.0 & - & - \\
        $\lambda/\epsilon_1/\epsilon_2$ for TNAR-LGAN (large) & 0.01/0.5/2.0 & 0.1/5.0/1.0 & - & - \\
        $\lambda/\epsilon_1/\epsilon_2$ for TNAR-VAE (small) & 0.01/0.2/2.0 & 0.01/5.0/1.0 & - & - \\
        $\lambda/\epsilon_1/\epsilon_2$ for TNAR-VAE (large) & 0.01/0.2/2.0 & 0.001/5.0/1.0 & 0.01/0.2/2.0 & 0.001/4.0/1.0 \\
        \hhline{=====}
    \end{tabular}
\end{table*}

\section{More adversarial examples}
\begin{figure}[ht]
\centering
\begin{tabular}{ccccc}
\includegraphics[width=0.14\linewidth]{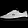}
& \includegraphics[width=0.14\linewidth]{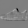}
& \includegraphics[width=0.14\linewidth]{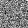}
& \includegraphics[width=0.14\linewidth]{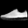}
& \includegraphics[width=0.14\linewidth]{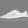}
\\ \includegraphics[width=0.14\linewidth]{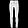}
& \includegraphics[width=0.14\linewidth]{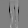}
& \includegraphics[width=0.14\linewidth]{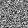}
& \includegraphics[width=0.14\linewidth]{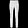}
& \includegraphics[width=0.14\linewidth]{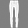}
\\ \includegraphics[width=0.14\linewidth]{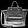}
& \includegraphics[width=0.14\linewidth]{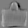}
& \includegraphics[width=0.14\linewidth]{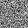}
& \includegraphics[width=0.14\linewidth]{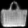}
& \includegraphics[width=0.14\linewidth]{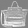}
\\ \includegraphics[width=0.14\linewidth]{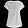}
& \includegraphics[width=0.14\linewidth]{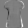}
& \includegraphics[width=0.14\linewidth]{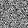}
& \includegraphics[width=0.14\linewidth]{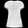}
& \includegraphics[width=0.14\linewidth]{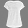}
\end{tabular}
\caption{\small The perturbations and adversarial examples in the tangent space and normal space for FashionMNIST dataset.
    Note that since the perturbations are actually too small, to distinguish them visually, thus we show the scaled perturbations.
From left to right: original example, tangent adversarial perturbation, normal adversarial perturbation,
tangent adversarial example, normal adversarial example.}
\label{fig:adversarial_fashion_more}
\end{figure}

\begin{figure}[ht]
\centering
\begin{tabular}{ccccc}
\includegraphics[width=0.14\linewidth]{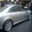}
& \includegraphics[width=0.14\linewidth]{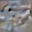}
& \includegraphics[width=0.14\linewidth]{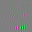}
& \includegraphics[width=0.14\linewidth]{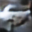}
& \includegraphics[width=0.14\linewidth]{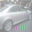}
\\ \includegraphics[width=0.14\linewidth]{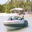}
& \includegraphics[width=0.14\linewidth]{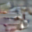}
& \includegraphics[width=0.14\linewidth]{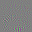}
& \includegraphics[width=0.14\linewidth]{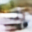}
& \includegraphics[width=0.14\linewidth]{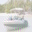}
\\ \includegraphics[width=0.14\linewidth]{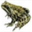}
& \includegraphics[width=0.14\linewidth]{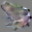}
& \includegraphics[width=0.14\linewidth]{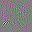}
& \includegraphics[width=0.14\linewidth]{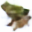}
& \includegraphics[width=0.14\linewidth]{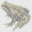}
\\ \includegraphics[width=0.14\linewidth]{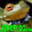}
& \includegraphics[width=0.14\linewidth]{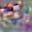}
& \includegraphics[width=0.14\linewidth]{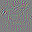}
& \includegraphics[width=0.14\linewidth]{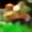}
& \includegraphics[width=0.14\linewidth]{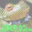}
\end{tabular}
\caption{\small The perturbations and adversarial examples in tangent space and normal space for CIFAR-10 dataset.
    Note that the perturbations is actually too small to distinguish easily, thus we show the scaled perturbations.
From left to right: original example, tangent adversarial perturbation, normal adversarial perturbation,
tangent adversarial example, normal adversarial example.}
\label{fig:adversarial_cifar_more}
\end{figure}

More adversarial perturbations and adversarial examples in the tangent space and normal space are shown in Figure~\ref{fig:adversarial_fashion_more} and Figure~\ref{fig:adversarial_cifar_more}.

\end{document}